\documentclass[conference]{IEEEtran}
\IEEEoverridecommandlockouts
\usepackage{hyperref}
\usepackage{cite}
\usepackage{amsmath,amssymb,amsfonts}
\usepackage{algorithmic}
\usepackage{graphicx}
\usepackage{textcomp}
\usepackage{cleveref}
\usepackage{xcolor}
\usepackage{xspace}
\usepackage{cite}
\usepackage{makecell}
\usepackage{float}

\def\BibTeX{{\rm B\kern-.05em{\sc i\kern-.025em b}\kern-.08em
    T\kern-.1667em\lower.7ex\hbox{E}\kern-.125emX}}
\begin{document}
\newcommand{\ie}{\textit{i}.\textit{e}., }
\newcommand{\R}{\mathbb{R}}
\newcommand{\yolovnine}{YOLOv9\xspace}
\newcommand{\regmax}{N}
\newcommand{\moe}{MoE\xspace} 

\title{Find the Leak, Fix the Split: Cluster-Based Method to Prevent Leakage in Video-Derived Datasets
\\
\thanks{*These authors contributed equally to this work.}
}

\author{
\makebox[\textwidth][c]{%
\begin{minipage}[t]{0.48\textwidth}
\centering
Noam Glazner*\\
\textit{Faculty of Engineering}\\
\textit{Bar-Ilan University}\\
Ramat Gan, Israel\\
Email: noam.glazner@live.biu.ac.il\\
ORCID: 0009-0003-1524-1157
\end{minipage}
\hfill
\begin{minipage}[t]{0.48\textwidth}\centering
Noam Tsfaty\\
\textit{Intelligence Systems}\\
\textit{Afeka College of Engineering}\\
Tel Aviv, Israel\\
Email: Tsfaty.Noam@s.afeka.ac.il\\
ORCID: 0009-0009-5246-8274
\end{minipage}

}\\[0.6em]
\makebox[\textwidth][c]{
\begin{minipage}[t]{0.48\textwidth}
\centering
Sharon Shalev\\
\textit{Independent Researcher}\\
Email: shalev1310@gmail.com \\
ORCID: 0009-0003-5054-230X
\end{minipage}
\hfill
\begin{minipage}[t]{0.48\textwidth}
\centering
Avishai Weizman*\\
\textit{School of Electrical and Computer Engineering}\\
\textit{Ben-Gurion University of the Negev}\\
Beersheba, Israel\\
Email: wavishay@post.bgu.ac.il \\
ORCID: 0009-0004-1182-8601
\end{minipage}
}%
}

\maketitle

\begin{abstract}
We propose a cluster-based frame selection strategy to mitigate information leakage in video-derived frames datasets. By grouping visually similar frames before splitting into training, validation, and test sets, the method produces more representative, balanced, and reliable dataset partitions.
\end{abstract}

\section{Introduction}
In deep learning research, database construction and data splitting strategies play a crucial role in determining the reliability and generalization capability of trained models. Image-based models can be trained by randomly splitting images into training, validation, and test sets~\cite{botache2023unravelingcomplexitysplittingsequential}.
However, when datasets are derived from video sources, this approach can lead to information leakage due to high spatial and temporal correlations among consecutive frames, e.g., frames recorded in the same background and containing the same objects but in slightly different positions or rotations~\cite{figueiredo2024analyzing,botache2023unravelingcomplexitysplittingsequential}. As a result, models can achieve inflated performance estimates, particularly in image-based object detection trained on video-derived frames.
To address this issue, we propose a cluster-based frame selection strategy that groups visually similar frames before splitting into the dataset partitions (train, validation, test), ensuring that correlated images remain together. The proposed method is simple, scalable, and can be integrated into existing dataset preparation pipelines without modification of the training process. 
This approach can improve the fairness of dataset partitioning and improve the robustness of model evaluation in video-derived frames datasets.
\section{Proposed Method}
\begin{figure}[t]
    \centering
    \includegraphics[width=1\linewidth]{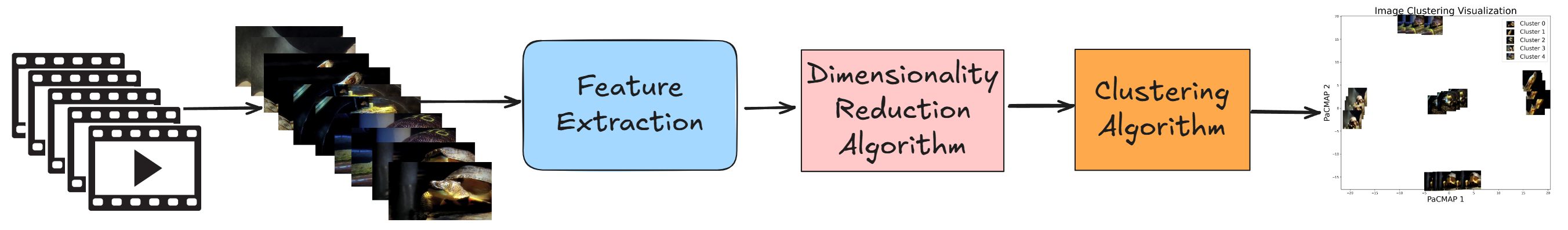}
    \caption{Illustration of the proposed cluster-based frame selection pipeline. Each video is decomposed into individual frames, from which features are extracted using semantic representations (e.g., CLIP~\cite{Radford2021LearningTV}), handcrafted descriptors (e.g., HOG~\cite{dalal2005histograms}), or lightweight learned features (e.g., XFeat~\cite{potje2024xfeat}). Dimensionality reduction (e.g., PaCMAP~\cite{wang2021understanding}) and clustering (e.g., HDBSCAN~\cite{mcinnes2017hdbscan}) are then applied to group visually similar frames before dataset partitioning.}
    \label{fig:overview}
\end{figure}

An overview of the proposed framework is presented in~\autoref{fig:overview}. 
The process begins by considering a collection of unlabeled videos denoted as $\mathcal{V} = \{V_1, V_2, \ldots, V_K\}$, where $K$ represents the total number of videos in the dataset. Each video $V_k$ is decomposed into a sequence of frames 
$\{I_{k,1}, I_{k,2}, \ldots, I_{k,N_k}\}$,
where $N_k$ denotes the number of frames extracted from video $V_k$.
\subsection{Feature Extraction}
For each frame $I_{k,i}$, a feature vector $\mathbf{f}_{k,i} \in \mathbb{R}^d$ is extracted using handcrafted or learned descriptors. Classical descriptors such as the \textit{scale-invariant feature transform} (SIFT)~\cite{lowe2004distinctive,790410} and the \textit{histogram of oriented gradients} (HOG)~\cite{dalal2005histograms} capture local texture and shape information by encoding gradient orientation patterns within localized image regions. 

The \textit{Accelerated Features for Lightweight Image Matching} (XFeat)~\cite{potje2024xfeat} descriptor provides efficient keypoint detection and description through a lightweight convolutional architecture, while deep pretrained models such as \textit{Contrastive Language-Image Pretraining} (CLIP)~\cite{Radford2021LearningTV}, \textit{Sigmoid Loss for Language Image Pre-training} (SigLIP)~\cite{zhai2023sigmoid}, and \textit{DINO-V3}~\cite{simeoni2025dinoV3} offer semantic image representations that can capture visual concepts.
The \textit{Vector of Locally Aggregated Descriptors} (VLAD)~\cite{jegou2010aggregating} was applied to both SIFT and XFeat to combine their local keypoint descriptors into a single compact, fixed-length feature vector. VLAD creates a compact representation by assigning each local descriptor to its nearest codeword in a learned codebook and aggregating the residuals between them into a single vector. The feature extraction process is defined as:

\begin{equation}
\mathbf{f}_{k,i} = \Phi_{\text{feat}}(I_{k,i}),
\end{equation}
where $\Phi_{\text{feat}}(\cdot)$ denotes the chosen feature extraction function.

\subsection{Dimensionality Reduction and Clustering}
The extracted feature vectors $\mathbf{f}_{k,i}$ are projected into a low-dimensional embedding space using the \textit{pairwise controlled manifold approximation projection} (PaCMAP) algorithm~\cite{wang2021understanding}:
\begin{equation}
\mathbf{z}_{k,i} = \mathcal{P}_{\text{PaCMAP}}(\mathbf{f}_{k,i})
\end{equation}
where $\mathbf{z}_{k,i} \in \mathbb{R}^m$  represents the embedded frame in the latent space, and $\mathcal{P}_{\text{PaCMAP}}(\cdot)$ denotes the projection operator applied by the PaCMAP algorithm.
Subsequently, the embedded representations ${\mathbf{z}_{k,i}}$ are clustered using the \textit{hierarchy of density-based spatial clustering} (HDBSCAN)~\cite{mcinnes2017hdbscan} algorithm.
Due to the continuous and non-uniform nature of video data, density-based clustering methods such as HDBSCAN are generally more suitable than centroid-based approaches (e.g., K-Means~\cite{ahmed2020k} algorithm), as they can capture irregular cluster shapes and variable frame densities, well-suited for this task.

\subsection{Cluster-Based Dataset Partitioning}
The HDBSCAN clustering algorithm is employed to group similar frames based on their feature representations $\mathbf{z}_{k,i}$. Let \(C_j\) denote the \(j\)-th cluster containing feature $\mathbf{z}_{k,i}$, which correspond to frame \(I_{k,i}\).
Each cluster \(C_j\) represents visually related frames, which are then assigned to the same dataset partition to prevent data leakage between sets.

\section{Experiments and Results}
\label{sec:Experiments_and_results}
We analyze the results in~\autoref{tab:imagenet_umap_pacmap_results} using the validation set of the ImageNet-VID set of the \textit{ImageNet Large Scale Visual Recognition Challenge 2015} (ILSVRC2015)~\cite{russakovsky2015imagenet} and all partitions of the UCF101~\cite{Soomro2012UCF101AD} dataset. The ImageNet dataset provides annotated images categorized by object synsets, whereas the UCF101 dataset consists of trimmed video clips labeled at the video level. 
To reduce visual redundancy and ensure that consecutive frames are not nearly identical, we extracted one frame per second from each video in the UCF101 dataset. 
These structured annotations allow us to detect and mitigate potential data leakage by ensuring that all frames from a given video remain within a single cluster.
In our experiments, the images were resized to $224{\times}224$ for XFeat, CLIP, DINO, and SigLIP, and to either $128{\times}128$ or $224{\times}224$ for HOG, with the $128{\times}128$ setting performing slightly better while also producing a lower-dimensional descriptor. The VLAD vectors were reduced to $1024$ dimensions to provide a uniform feature representation, and the PaCMAP embeddings $\mathbf{z}_{k,i}$ were projected to a 256-dimensional space ($m = 256$) to ensure a consistent low-dimensional representation across all feature extraction methods.
To evaluate the effectiveness of the proposed frame grouping method, we assess clustering quality using the \textit{Adjusted Mutual Information} (AMI)~\cite{vinh2009information} and \textit{V-measure}~\cite{Rosenberg2007VMeasureAC} metrics, which quantify how well the resulting clusters correspond to the ground-truth videos. 
The AMI measures the agreement between predicted clusters and true labels while accounting for chance, whereas the V-measure evaluates the trade-off between cluster homogeneity (the extent to which each cluster is composed of samples from a single class) and completeness (the extent to which samples of the same class share a cluster).
\begin{table}[t]
\centering
\caption{Clustering performance of HDBSCAN on validation sets from the ImageNet-VID and UCF101 datasets using different feature extractions and PaCMAP (256-D).}
\label{tab:imagenet_umap_pacmap_results}
\begin{tabular}{|c|c|c|c|}
\hline
\textbf{Feature Extraction} & \textbf{Dataset} & \textbf{V-measure} & \textbf{AMI} \\
\hline
SIFT + VLAD & ImageNet-VID & 0.81 & 0.80 \\
    & UCF101       & 0.57 &  0.38 \\
\hline
HOG & ImageNet-VID & 0.82 & 0.81 \\
    & UCF101       & 0.61 & 0.48 \\
\hline
HOG ($128 \times 128 $)& ImageNet-VID & 0.87 & 0.86 \\
    & UCF101       & 0.67 & 0.54 \\
\hline
XFeat + VLAD  & ImageNet-VID & 0.90 & 0.89 \\
      & UCF101       & 0.72 & 0.58 \\
\hline
CLIP (ViT-B/32) & ImageNet-VID & 0.92 & 0.91 \\
                 & UCF101       & 0.75 & 0.66 \\
\hline
SigLIP (ViT-B/16) & ImageNet-VID &0.93 & 0.92\\
                 & UCF101       &0.75&0.67\\
\hline
DINO-V3 (ViT-B/16) & ImageNet-VID &0.96& 0.96 \\
                 & UCF101      &0.87&0.80 \\
\hline
\end{tabular}
\end{table}
In \autoref{tab:imagenet_umap_pacmap_results}, all feature extraction methods use an input image size of $224 \times 224$ except for HOG ($128 \times 128$). The results show that representations from deep pretrained models outperform classical and lightweight descriptors on both datasets, demonstrating their superior ability to capture the similarity of information leakage detection. In particular, DINO-V3 achieves the highest V-measure and AMI scores on both ImageNet-VID and UCF101. XFeat, combined with VLAD, also provides a noticeable improvement over traditional descriptors, while SIFT + VLAD remains the weakest across both datasets, especially on UCF101, where temporal variability increases the difficulty of clustering.

\section{Conclusions and Future Work}
\label{sec:Conclusions_and_future_work}
This work presents a simple cluster-based frame selection strategy designed to minimize information leakage in video datasets. By grouping visually similar and temporally correlated frames before dataset partitioning, the proposed method can improve diversity and provide a fair evaluation setting, thereby mitigating overfitting in object detection models trained on video datasets.
The experimental results demonstrate that DINO-V3 embeddings achieve the best clustering across both ImageNet-VID and UCF101 datasets, highlighting the advantage of representations for identifying information leakage.
One limitation of the approach is its reliance on HDBSCAN and its hyperparameter choices, and we plan to explore adaptive clustering strategies that may address this limitation. We also aim to quantify the performance gap in image-based object detection models by training with and without the proposed approach to assess the impact of information leakage across different datasets.

\bibliographystyle{IEEEtran}
\bibliography{references}

@String(CVPR  = {IEEE Conf. Comput. Vis. Pattern Recog.})

@String(CVPR  = {CVPR})

@article{russakovsky2015imagenet,
  title={Imagenet large scale visual recognition challenge},
  author={Russakovsky, Olga and Deng, Jia and Su, Hao and Krause, Jonathan and Satheesh, Sanjeev and Ma, Sean and Huang, Zhiheng and Karpathy, Andrej and Khosla, Aditya and Bernstein, Michael and others},
  journal={International journal of computer vision},
  volume={115},
  number={3},
  pages={211--252},
  year={2015},
  publisher={Springer}
}

@article{lowe2004distinctive,
  title={Distinctive image features from scale-invariant keypoints},
  author={Lowe, David G},
  journal={International journal of computer vision},
  volume={60},
  number={2},
  pages={91--110},
  year={2004},
  publisher={Springer}
}

@inproceedings{zhai2023sigmoid,
  title={Sigmoid loss for language image pre-training},
  author={Zhai, Xiaohua and Mustafa, Basil and Kolesnikov, Alexander and Beyer, Lucas},
  booktitle={Proceedings of the IEEE/CVF international conference on computer vision},
  pages={11975--11986},
  year={2023}
}

@misc{botache2023unravelingcomplexitysplittingsequential,
      title={Unraveling the Complexity of Splitting Sequential Data: Tackling Challenges in Video and Time Series Analysis}, 
      author={Diego Botache and Kristina Dingel and Rico Huhnstock and Arno Ehresmann and Bernhard Sick},
      year={2023},
      eprint={2307.14294},
      archivePrefix={arXiv},
}

@inproceedings{Radford2021LearningTV,
  title={Learning Transferable Visual Models From Natural Language Supervision},
  author={Alec Radford and Jong Wook Kim and Chris Hallacy and Aditya Ramesh and Gabriel Goh and Sandhini Agarwal and Girish Sastry and Amanda Askell and Pamela Mishkin and Jack Clark and Gretchen Krueger and Ilya Sutskever},
  booktitle={International Conference on Machine Learning},
  year={2021},
}

@article{ahmed2020k,
  title={The k-means algorithm: A comprehensive survey and performance evaluation},
  author={Ahmed, Mohiuddin and Seraj, Raihan and Islam, Syed Mohammed Shamsul},
  journal={Electronics},
  volume={9},
  number={8},
  pages={1295},
  year={2020},
  publisher={MDPI}
}

@ARTICLE{figueiredo2024analyzing,
  author={Figueiredo, Ravi B. D. and Mendes, Hugo A.},
  journal={IEEE Access}, 
  title={Analyzing Information Leakage on Video Object Detection Datasets by Splitting Images Into Clusters With High Spatiotemporal Correlation}, 
  year={2024},
  volume={12},
  pages={47646-47655},
  doi={10.1109/ACCESS.2024.3383047}
}

@inproceedings{Rosenberg2007VMeasureAC,
  title={V-Measure: A Conditional Entropy-Based External Cluster Evaluation Measure},
  author={Andrew Rosenberg and Julia Hirschberg},
  booktitle={Conference on Empirical Methods in Natural Language Processing},
  year={2007}
}

@article{simeoni2025dinov3,
  title={Dinov3},
  author={Sim{\'e}oni, Oriane and Vo, Huy V and Seitzer, Maximilian and Baldassarre, Federico and Oquab, Maxime and Jose, Cijo and Khalidov, Vasil and Szafraniec, Marc and Yi, Seungeun and Ramamonjisoa, Micha{\"e}l and others},
  journal={arXiv preprint arXiv:2508.10104},
  year={2025}
}

@inproceedings{vinh2009information,
  title={Information theoretic measures for clusterings comparison: is a correction for chance necessary?},
  author={Vinh, Nguyen Xuan and Epps, Julien and Bailey, James},
  booktitle={Proceedings of the 26th annual international conference on machine learning},
  pages={1073--1080},
  year={2009}
}

@INPROCEEDINGS{jegou2010aggregating,
  author={Jégou, Hervé and Douze, Matthijs and Schmid, Cordelia and Pérez, Patrick},
  booktitle={2010 IEEE Computer Society Conference on Computer Vision and Pattern Recognition}, 
  title={Aggregating local descriptors into a compact image representation}, 
  year={2010},
  volume={},
  number={},
  pages={3304-3311},
  doi={10.1109/CVPR.2010.5540039}}

@inproceedings{dalal2005histograms,
  title={Histograms of oriented gradients for human detection},
  author={Dalal, Navneet and Triggs, Bill},
  booktitle={2005 IEEE computer society conference on computer vision and pattern recognition (CVPR'05)},
  volume={1},
  pages={886--893},
  year={2005},
  organization={Ieee}
}

@INPROCEEDINGS{790410,
  author={Lowe, D.G.},
  booktitle={Proceedings of the Seventh IEEE International Conference on Computer Vision}, 
  title={Object recognition from local scale-invariant features}, 
  year={1999},
  volume={2},
  number={},
  pages={1150-1157 vol.2},
  doi={10.1109/ICCV.1999.790410}
}

@article{Soomro2012UCF101AD,
  title={UCF101: A Dataset of 101 Human Actions Classes From Videos in The Wild},
  author={Khurram Soomro and Amir Zamir and Mubarak Shah},
  journal={ArXiv},
  year={2012},
  volume={abs/1212.0402},
}

@inproceedings{potje2024xfeat,
  author       = {Guilherme A. Potje and
                  Felipe Cadar and
                  Andr{\'{e}} Ara{\'{u}}jo and
                  Renato Martins and
                  Erickson R. Nascimento},
  title        = {XFeat: Accelerated Features for Lightweight Image Matching},
  booktitle    = {{IEEE/CVF} Conference on Computer Vision and Pattern Recognition
                  {CVPR} , June 16-22, 2024},
  pages        = {2682--2691},
  publisher    = {{IEEE}},
  year         = {2024},
  doi          = {10.1109/CVPR52733.2024.00259},
}

@article{wang2021understanding,
  author  = {Yingfan Wang and Haiyang Huang and Cynthia Rudin and Yaron Shaposhnik},
  title   = {Understanding How Dimension Reduction Tools Work: An Empirical Approach to Deciphering t-SNE, UMAP, TriMap, and PaCMAP for Data Visualization},
  journal = {Journal of Machine Learning Research},
  year    = {2021},
  volume  = {22},
  number  = {201},
  pages   = {1--73},
}

@article{mcinnes2017hdbscan,
  title={hdbscan: Hierarchical density based clustering.},
  author={McInnes, Leland and Healy, John and Astels, Steve and others},
  journal={J. Open Source Softw.},
  volume={2},
  number={11},
  pages={205},
  year={2017}
}

\end{document}